\def\year{2021}\relax
\documentclass[letterpaper]{article} 
\usepackage{aaai21}  
\usepackage{times}  
\usepackage{helvet} 
\usepackage{courier}  
\usepackage[hyphens]{url}  
\usepackage{graphicx} 
\usepackage{amsmath}
\usepackage{natbib}  
\usepackage{caption} 
\usepackage{booktabs, multirow} 
\usepackage{soul}
\usepackage[table]{xcolor} 
\usepackage{changepage,threeparttable} 

\title{Two-Sided Fairness in Non-Personalised Recommendations}
\author{

    Aadi Swadipto Mondal\thanks{The first three authors have contributed equally.},
    Rakesh Bal\textsuperscript{\rm $*$},
    Sayan Sinha\textsuperscript{\rm $*$},
    Gourab K Patro\\
}
\affiliations{

    \textsuperscript{\rm }Indian Institute of Technology Kharagpur, India - 721302.\\

    \{aadismondal, rakesh.bal, sayan.sinha, patrogourab\}@iitkgp.ac.in

}

\begin{document}

\maketitle

\begin{abstract}
\vspace{-2mm}
Recommender systems are one of the most widely used services on several online platforms to suggest potential items to the end-users.
These services often use different machine learning techniques for which fairness is a concerning factor, especially when the downstream services have the ability to cause social ramifications.
Thus, focusing on the non-personalised (global) recommendations in news media platforms (e.g., top-$k$ trending topics on Twitter, top-$k$ news on a news platform, etc.), we discuss on two specific fairness concerns together (traditionally studied separately)---user fairness \cite{chakraborty2019equality} and organisational fairness \cite{burke2020algorithmic}. 
While user fairness captures the idea of representing the choices of all the individual users in the case of global recommendations, organisational fairness tries to ensure politically/ideologically balanced recommendation sets.
This makes user fairness a user-side requirement and organisational fairness a platform-side requirement.
For user fairness, we test with methods from social choice theory, i.e., various voting rules known to better represent user choices in their results.
Even in our application of voting rules to the recommendation setup, we observe high user satisfaction scores (table \ref{tab:result}).
Now for organisational fairness, we propose a bias metric (eq-\ref{eq:bias}) which measures the aggregate ideological bias of a recommended set of items (articles).
Analysing the results obtained from voting rule-based recommendation, we find that while the well-known voting rules are better from the user side, they show high bias values and clearly not suitable for organisational requirements of the platforms.
Thus, there is a need to build an encompassing mechanism by cohesively bridging ideas of user fairness and organisational fairness.
In this abstract paper, we intend to frame the elementary ideas along with the clear motivation behind the requirement of such a mechanism.
\end{abstract}


\begin{table*}[!htp]\centering
\caption{Some of the seed words having high PMI values}\label{tab:word}
\scriptsize
\vspace{-3mm}
\begin{tabular}{lrr}\toprule
\textbf{Left-biased} &motstandernes, djevlene, motstander, problem, motstand, scorer, møte, sjef, kontret, balanse, maktet, rød, igjen \\
\textbf{Right-biased} &bakgrunnen, grunn, etablerte, opprettelsen, døpt, alternativ, offensivt, defensiv, renter, tilværelsen, høyre, erna, siv \\
\bottomrule
\vspace{-5mm}
\end{tabular}
\vspace{-5mm}
\end{table*}
\vspace{-5mm}
\section{Two-Sided Fairness}
\subsubsection{Dataset} The dataset used for this research is a publicly available news dataset that includes news articles (in Norwegian) in connection with anonymised users \cite{gulla2017adressa}. 
From this dataset, we use the 865 articles with publicly available news contents and the news consumption history (how much time a user spends reading different articles) of 63260 anonymised users.

\subsubsection{Non-personalised Recommendation}
Let $\mathcal{U}$ and $\mathcal{A}$ be the set of users and articles (items) respectively.
Most of the recommender systems rely on some form of scoring for user-item pairs (e.g., ratings).
Here we derive such scores from the ``Active Time" of a user on an article (the total amount of time in seconds the user spends on the article). 
We derive the score for user $u\in \mathcal{U}$ and article $a\in \mathcal{A}$ pair as:
{$Score(u, a)= \dfrac{ActTime(u, a)}{\sum_{u'\in \mathcal{U}} ActTime(u', a)}\cdot N_a$}.    
Here $ActTime(u, a)$ is the total active time of the user $u$ on article $a$ and $N_a$ is the number of users who have viewed the article at least once. 
Next, we linearly scale all the scores of each user separately to floating-point numbers in the range (1 to 10), in order to ensure similar scales for fast and slow readers.
This choice of scores is purely dataset-specific and can be replaced with other suitable score metric as per the requirement.
Here, a higher score signifies more affinity of a user towards the article. However, there are many user-article pairs for which a score does not exist, owing to the fact that the user might not have viewed that particular article. 
We find these missing values using Non-Negative Matrix Factorisation (NMF) \cite{koren2009matrix}.

\noindent {\bf Non-negative Matrix Factorisation (NMF):}
We obtain a sparse matrix $V$ from the normalised scaled scores obtained earlier. 
The rows of $V$ represent the users, and the columns represent the articles. 
We then perform NMF on $V$ and obtain $\hat V$ which provides us with the predicted scores for the missing values in $V$. $V^*$ denotes the final score matrix, which comprises the original scores from $V$, as well as the predicted scores from $\hat V$ (for the missing entries in $V$).
NMF was set on training with hyperparameters $\alpha = 0.0002$, and $\beta = 0.02$. Convergence was assumed when $\Delta cost < 0.001$ (around $2,500$ epochs).

\noindent {\bf The Recommendation:} As we talk about non-personalised recommendation, the goal is to find $\kappa$-sized recommendation which would be liked by most of the users. For this purpose, it is wiser to use $V^*$ as it---being a complete score matrix---can provide insights on all the user-article pairs. Next, we discuss two kinds of fairness desirable in such recommendations and also the obstacles in achieving them.

\vspace{-1mm}
\subsubsection{User Fairness}
As the recommendation is global, it should fairly represent the interests of the users on the platform.
Thus, we can re-purpose and use multi-winner voting rules in our setup as voting rules are known to fairly represent voters' choices while choosing winners.
To map our setup to an election we do the following;
election $\mathcal{E} = (\mathcal{C}, \mathcal{V})$, where candidate set $\mathcal{C}=\mathcal{A}= (a_{1}, . . . , a_{m})$ is the set of articles, voter set $\mathcal{V}=\mathcal{U} = (u_{1}, . . . , u_{n})$ is the set of users, and we consider scores $V^*$ as the corresponding votes.
The winners of the election are denoted by the set $\mathcal{W} = (w_{1}, . . . , w_{\kappa})$ which is a subset of $\mathcal{C}$ and are elected using various voting rules.
In this work, we experimented with six famous voting rules, namely SNTV, STV, $k$-Borda, Bloc voting, Chamberlin Courant and Monroe \cite{elkind2017properties}. 
Next, we find the average satisfaction of all users for the results obtained from each election method like:
$ Satisfaction$
$=$
$\dfrac{1}{\mathcal{U}} \sum_{u\in \mathcal{U}} \frac{ \mathcal{W} \cap Top\text{-}\kappa_u}{\kappa}$.
Here $Top\text{-}\kappa_u$ is the top $\kappa$ articles for user $u$ as per the scores in $V^*$. The satisfaction score for each election method has been presented in Table \ref{tab:result} column 2. A high satisfaction score would ensure that the recommendation results (election winners) hence obtained follow the general principle of user fairness. In all our experiments, the value of $\kappa$ has been taken as 10.
\vspace{-1mm}
\subsubsection{Organisational Fairness}
For organisational fairness \cite{burke2020algorithmic}, the set of recommended articles needs to be balanced in terms of political/ideological bias.
Before finding political bias of the articles in our dataset,
first, we analyse articles from politically-biased media houses -- namely, \textit{Klassekampen} (left) and \textit{Aftenposten} (right) \cite{Norwegia50:online}, as well as articles talking about eminent people involved in politics pertaining to either of these ideologies.
Using co-occurrence analysis of words in the articles of biased media houses, we shortlist some of the words with the highest Pointwise Mutual Information (PMI) scores. 
Some of the words having high PMI values---which we later use as seed words---have been mentioned in Table \ref{tab:word}.
We consider that their presence in an article indicates a higher likelihood of it being politically biased.
Thus, we designate the articles in our dataset as either left-biased or right-biased based on whether they contain the majority of seed words from the left category or right category.
Next, we find the reference bias ($\rho$) of the user population on the platform as
the ratio of total time spent on the left-biased articles to that on the right-biased articles\footnote{In our experiments, $\rho$ was found to be $1.423$.}. 
Finally, we design a bias metric \footnote{More sophisticated methods for bias analysis can also be used with the availability of relevant annotated resources in Norwegian. However this is not under the purview of this abstract.} which measures how far are the global recommendations from reference $\rho$.
\begin{equation}\label{eq:bias}
    Bias = 
    \dfrac{-leftCount + \rho * rightCount}{leftCount + \rho * rightCount}
\end{equation}
Here $leftCount$ and $rightCount$ are the number of occurrences of left and right biased seed words in a set of articles. Hence, $Bias<0$ indicates left-bias and $Bias>0$ indicates right-bias. 
Closer the value of $Bias$ to zero, better is the method in terms of satisfying organisation fairness. The range of $Bias$ is [-1, 1]. 
We calculated the bias values of the global recommendations obtained earlier using voting rules.
The results have been shown in Table \ref{tab:result} column 3. 

\begin{table}[!htp]\centering
\tiny
\caption{User Satisfaction and Organisational Bias}\label{tab:result}
\scriptsize
\vspace{-3mm}
\begin{tabular}{lrrr}\toprule
\textbf{Election Method} &\textbf{Satisfaction} &\textbf{Bias} \\\midrule
SNTV &0.878 &-0.115 \\
$k$-Borda &0.800 &0.033 \\
Bloc &0.856 &-0.240 \\
STV &0.894 &-0.117 \\
CC &0.883 &-0.169 \\
Monroe &0.825 &-0.103 \\
\bottomrule
\end{tabular}
\end{table}
\subsubsection{Discussion}
For better user fairness, we repurposed well-known voting rules from fair voting theory, and implemented the same in our recommendation setup.
Our evaluation shows that they achieve good user satisfaction scores (column 2 of table \ref{tab:result}).
However, they show high bias values (column 3 of table \ref{tab:result}).
This is happening as the voting rules do not consider the article content while choosing the global recommendations.
On the other hand, organisational fairness---relating to balance in the content distribution---requires the bias to be close to zero.
Therefore, there is a need for designing a mechanism which simultaneously cares for user fairness and organisational fairness while designing non-personalised recommendations. All of our resources can be found at \url{https://github.com/americast/electoral}. 

\subsubsection{Acknowledgement} 
We would like to thank Prof Niloy Ganguly (CSE, IIT Kharagpur) and Rishabh Joshi (LTI, CMU) for their continued support throughout the development of this work.
Gourab K Patro is supported by a fellowship from Tata Consultancy Services.
{\small\bibliography{aaai21}}
\end{document}